\documentclass[conference]{IEEEtran}
\IEEEoverridecommandlockouts
\usepackage{cite}
\usepackage{amsmath,amssymb,amsfonts}
\usepackage{algorithmic}
\usepackage{graphicx}
\usepackage{textcomp}
\usepackage{xcolor}
\usepackage{xspace}
\usepackage{url}

\def\BibTeX{{\rm B\kern-.05em{\sc i\kern-.025em b}\kern-.08em
    T\kern-.1667em\lower.7ex\hbox{E}\kern-.125emX}}
\begin{document}

\newcommand\iuxray{\textsc{IUXRay}\xspace}
\newcommand\mimic{\textsc{MIMIC-III}\xspace}
\newcommand\mimicr{\textsc{MIMIC-III@radio}\xspace}
\newcommand\mimicd{\textsc{MIMIC-III@disch}\xspace}

\title{Clinical Predictive Keyboard using Statistical and Neural Language Modeling}

\author{\IEEEauthorblockN{John Pavlopoulos}
\IEEEauthorblockA{\textit{Department of Computer and System Sciences} \\
\textit{Stockholm University, Sweden}\\
ioannis@dsv.su.se}
\and
\IEEEauthorblockN{Panagiotis Papapetrou}
\IEEEauthorblockA{\textit{Department of Computer and System Sciences} \\
\textit{Stockholm University, Sweden}\\
panagiotis@dsv.su.se}
}

\maketitle

\begin{abstract}
A language model can be used to predict the next word during authoring, to correct spelling or to accelerate writing (e.g., in sms or emails). Language models, however, have only been applied in a very small scale to assist physicians during authoring (e.g., discharge summaries or radiology reports). But along with the assistance to the physician, computer-based systems which expedite the patient's exit also assist in decreasing the hospital infections. We employed statistical and neural language modeling to predict the next word of a clinical text and assess all the models in terms of accuracy and keystroke discount in two datasets with radiology reports. We show that a neural language model can achieve as high as 51.3\% accuracy in radiology reports (one out of two words predicted correctly). We also show that even when the models are employed only for frequent words, the physician can save valuable time.
\end{abstract}


\section{Introduction}

Text prediction is a challenging problem in machine learning and natural language processing, while at the same time there is a growing need for novel techniques for efficient and accurate text prediction in several application domains, such as in dictation and typing systems for people with disabilities or clinical text prediction for healthcare practitioners \cite{10.1007/s10209-005-0005-9}. More concretely, with text prediction we refer to the task of predicting the next block of text in an online fashion, where block can refer to different text granularity levels, e.g., sentences, words, syllables, or characters (keystrokes) \cite{10.1155/2016/3054258}.

The main focus of this paper is medical text with the concrete task of predicting the next word given an incomplete text. We also refer to this problem as \emph{predictive keyboard} for medical text. When applied in the clinical setting (e.g., authoring of hospital discharge summaries or diagnostic text), physicians can vastly benefit from a fast and accurate predictive keyboard system, since it can assist them with (a) a speedy compilation of the intended text, (b) means for prevention of potential text errors due to work overload, (c) means for speedier patient discharge.

Initial efforts towards solving the predictive keyboard problem for radiology reports are described by Eng and Eisner 2004 \cite{Eng2004}, where a 3-Gram language model achieves an average keystroke reduction of a factor of 3.3. Following this line of research, we employed N-Gram-based statistical language modeling, which refer to as N-GLM, to predict the next word of a clinical text. We vary N from 1 to 10 and show that 4-Gram models achieve 38\% accuracy when predicting the next word in a clinical text, outperforming other N-GLMs. Observe that accuracy in this case measures the fraction of times when the next word was predicted correctly, hence inducing an equivalent typing speedup at the word level. We additionally investigated two neural language models that employ (1) a Recurrent Neural Network (RNN) language model based on Long-Short Term Memory, which we refer to as LSTMLM \cite{Hochreiter1997} and (2) a Gated Recurrent Unit (GRU) based language model, which we refer to as GRULM. This model achieves higher levels of accuracy compared to 4-GLM, since our experimental evaluation demonstrates that accuracy can reach up to 51.3\% (i.e., 5 out of 10 `next' words predicted correctly). An example of the output of this task is depicted in Table \ref{tab:my_label}, where we can observe the next word predictions made by LSTMLM and 4-GLM, with the correctly predicted words indicated in '[]'.

Next, we outline the related work in the area of clinical text prediction, followed by a summary of our contributions.

\begin{table}[!t]
    \centering
    \caption{Example use-case on \iuxray test words, using 4-GLM and LSTMLM. Words in [] were correctly predicted by each model.}
    \begin{tabular}{|c|p{7cm}|}\hline
         LSTMLM & "the lungs are clear without [evidence] [of] focal infiltrate [or] [effusion] [there] [is] [no] [pneumothorax] [the] [visualized] [bony] [structures] [reveal] [no] [acute] [abnormalities]"   \\\hline
        4-GLM  & "the lungs are [clear] without evidence [of] [focal] infiltrate or effusion [there] [is] [no] pneumothorax [the] visualized bony [structures] reveal [no] [acute] [abnormalities]"\\\hline 
    \end{tabular}
    \label{tab:my_label}
\end{table}

\subsection{Related Work}
The study of the benefits of computer-assisted text generation dates back to more than two decades ago \cite{Koester1994}. When applied to clinical notes, such as radiology reports, a statistical 3-Gram language model (including back off) achieved substantial keystroke reductions \cite{Eng2004}. Recently, an even simpler 3-Gram language model (i.e., with no back off) outperformed the earlier 
while also decreasing the typing time for the clinician by one third \cite{Yazdani2019a}. These results demonstrate that N-Gram models can provide promising solutions to our problem, and hence in this paper we provide a more extensive evaluation of these models on medical text.
Besides computer-assisted typing, language models have also been used for spelling correction in clinical notes \cite{Patrick2010,Yazdani2019b}. This work does not focus on spelling correction, but what these works verify is that the words suggested by the language model during typing, are also checked for their correctness (i.e., assuming that the corpus contains correct words), hence the generated text will be of equal or even higher quality. 

With the recent advance of deep learning, deep neural networks, such as Long-short Term Memory (LSTM) \cite{Hochreiter1997} models, have improved the performance in natural language processing (NLP) tasks of the biomedical field, such as Name Entity Recognition (NER) \cite{Lee2020}; medical codes prediction \cite{Mullenbach2018}; relation classification \cite{Luo2017}; predicting hospital readmission \cite{Huang2019}. And language modeling is also part of this advent, since it is often employed as a pre-training step \cite{Radford2019}. For the task of next word prediction in a medical setting, however, neural language modeling is heavily under explored. To our knowledge, the only application of a neural language model was that of a baseline LSTM network (applied on a private dataset), which was improved when structured information from electronic health data (e.g., gender or age) was integrated \cite{Spithourakis2016}. The authors reported 8\% Accuracy (a.k.a. Recall@1 or Precision@1) for the baseline LSTM, which ranks it much lower than competing statistical language models \cite{Luo2017}. However, neural networks have been reported to outperform statistical language modeling in non-medical domains \cite{Dauphin2017}. 
In this work, we compare statistical and neural language modeling, a comparison which has not been studied before, and we show that the neural approach outperforms the statistical approach in next word prediction by a large margin.

\subsection{Contributions}
The main contributions of this paper can be summarized as follows: (1) We highlight the importance of the problem of keyword prediction for clinical text, and demonstrate how language models can be employed for providing scalable solutions to this problem; (2) We provide an extensive benchmark on clinical text obtained from two real-world medical datasets by comparing the performance of the N-GLM model for different values of N in terms of accuracy
and keystroke reduction;
(3) We additionally compare an RNN language model based on LSTM and GRU on the same datasets and demonstrate their superiority against N-GLMs as they can achieve an accuracy of up to 51.3\%,
indicating a speedup (at the word level) of the same degree, and a keyword reduction of up to 41.12\%, indicating a speedup (at the character level) of the same degree.

\section{Methods}
\label{sec:methods}

\subsection{Statistical Language Modeling}
Statistical language models \cite{Jelinek1997,Jurafsky2000} are based on the Markov assumption, modeling the probability of the next word, but given only the $n-1$ preceding words. The counts of all sequences of $n$ words (a.k.a. $n$-grams) are calculated over a corpus and a probability distribution over the vocabulary is modeled for each gram of $n-1$ words:
\begin{equation}
\label{eq:slm_probs}
    P(w_i | w_{i-1}, \dots, w_1) = P(w_i | w_{i-1},\dots,w_{i-n+1}) 
\end{equation}

Then, a $2$-gram (a.k.a. bigram) model will only consider the previous word $w_{i-1}$ to predict a next word $w_i$. And $w_i$ will be the one most frequently occurring in the corpus right after $w_{i-1}$. Probabilities are formed using the maximum likelihood estimation, changing Eq.~\ref{eq:slm_probs} to:
\begin{equation}
\label{eq:slm_counts}
    P(w_i | w_{i-1}) = \frac{C(w_{i-1}w_i)}{C_{i-1}}
\end{equation}

\noindent
where $C$ are the counts of the gram. To deal with unknown words, a pseudo token can be introduced (e.g., masking very rare words with `\textsc{[oov]}' during training). And to deal with unseen sequences of words one can introduce smoothing or backoff and interpolation. In this work we employed Laplace smoothing, but we observe that algorithms such as the Knesser-Ney or the Good-Turing backoff should also be investigated. For more information regarding statistical language models we redirect the interested reader to \cite{Jurafsky2000}.

\subsection{Neural Language Modeling}
Neural language modeling makes it possible to consider long range word dependencies without an explicitly predefined context length \cite{Sundermeyer2015}. The neural language model at each time step $s$ learns a hidden state $h_{s}$ as the non-linear combination (the weight matrix $W$ is learned) of the input word $x_{s}$ and the previous hidden state $h_{s-1}$. The vanishing gradient problem, arising from the deep in time back-propagation, is addressed with the LSTM cells \cite{Hochreiter1997}. More formally:

\begin{equation}
\begin{split}
    i_{s} &= \sigma(W_{i} \cdot [x_s, h_{s-1}] + b_{i})\\
    f_{s} &= \sigma(W_{f} \cdot [x_{s}, h_{s-1}] + b_{f})\\
    o_{s} &= \sigma(W_{o} \cdot [x_{s}, h_{s-1}] + b_{o})\\
    q_{s} &= \tanh(W_{q} \cdot [x_{s}, h_{s-1}] + b_{q})\\
    c_{s} &= f_{s} \cdot c_{s-1} + i_{s} \cdot q_{s}\\
    h_{s} &= o_{s} \cdot \tanh(c_{s}),
\end{split}
\end{equation}

\noindent
where $i_{s}$ is the input gate and $f_{s}$ is the forget get, which regulate the information from this ($q_{s}$) and the previous ($c_{s-1}$) cell to be forgotten, and $o_{s}$ is the output gate which regulates the information of the new hidden state. Then, the generation of the next word $x_{s+1}$ can be seen as a classification task, with $softmax$ yielding a probability distribution over the whole vocabulary and the next word to be generated being the most probable one.

In this work we also experiment with a different RNN variant, called Gated Recurrent Unit (GRU) \cite{Cho2014}, which is considered to be more efficient than LSTMs \cite{Chung2014}. It has a similar formulation with LSTMs:
\begin{equation}
\begin{split}
    r_{s} &= \sigma(W_{i} \cdot [x_s, h_{s-1}] + b_{i})\\
    u_{s} &= \sigma(W_{f} \cdot [x_{s}, h_{s-1}] + b_{f})\\
    c_{s} &= \tanh(W_{c} \cdot [x_{s}, r_{s} \cdot h_{s-1}] + b_{c})\\
    h_{s} &= u_{s} \cdot c_{s-1} + 1-u_{s} \cdot c_{s}\\
\end{split}
\end{equation}

\noindent
where $r_{s}$ and $u_{s}$ are the reset and update gates, defined similarly to the input and forget LSTM gates. No output gate is used, leading to a smaller number of gates and less computations, which makes GRU more efficient than LSTM.

\section{Empirical Evaluation}

\subsection{Datasets}
\label{sec:data}
We used two real-world medical datasets.

\smallskip
\noindent
\textbf{\iuxray.} The dataset comprises 3,955 anonymized and de-identified radiology reports on 7,470 images \cite{DF2015}\footnote{\url{https://openi.nlm.nih.gov/}}. The text of each report follows an XML structure and the boundaries of each different section are explicitly defined. 

\smallskip 
\noindent
\textbf{\mimic.} We used the radiology reports from the Medical Information Mart for Intensive Care (\mimic) \cite{Johnson2016} database, a rich and commonly used benchmark dataset of 38,597 adult patients admitted between 2001 and 2008 to critical care units at Beth Israel Deaconess Medical Center in Boston, Massachusetts. In this study we employ the free text reports of electrocardiogram and imaging studies included in this dataset. The text of the radiology reports in \mimic is loosely separated in sections, which are not explicitly marked up. We sampled 2,928 such reports to yield a dataset equal in number to \iuxray.

\smallskip
\noindent
The radiology reports of \iuxray and \mimic comprise less than 200 tokens per report in average.  By contrast, the discharge summaries are lengthier and more than quadruple in size. The difference grows larger when sampling disregards the maximum number of characters per text, because only discharge summaries did exceed this threshold.\footnote{The average number of words per summary without sampling is 1320.}

\begin{table}[!t]
    \centering
    \caption{Assessment of next word prediction in the radiology reports of \iuxray and \mimic, using statistical (N-GLMs) and neural (LSTMLM, GRULM) language models. Micro-averaged accuracy (Acc) and keystroke discount (KD) are shown for each dataset.}
    \begin{tabular}{c|c|c|c|c}
    & \multicolumn{2}{|c|}{\iuxray} & \multicolumn{2}{|c}{\mimic} \\
    & \sc Acc & \sc KD & \sc Acc & \sc KD\\\hline
2-GLM & 21.83±0.29 & 16.04±0.26 & 17.03±0.22 &11.46±0.12\\
3-GLM & 34.78±0.38 & 27.96±0.27& 27.34±0.29 &19.35±0.27\\
4-GLM & 38.18±0.44 &31.60±0.30& 25.70±0.29 &18.95±0.34\\
5-GLM & 37.89±0.60 &32.30±0.47& 21.02±0.41 &15.63±0.23\\
6-GLM & 35.71±0.78 &30.86±0.57& 15.98±0.42 &11.93±0.31\\
7-GLM & 33.10±0.72 &28.82±0.56& 12.15±0.40 &9.05±0.26\\
8-GLM & 30.23±0.63 &26.47±0.62& 9.52±0.40 &7.04±0.31\\
9-GLM & 27.74±0.63 &24.33±0.66& 7.29±0.43 &5.46±0.37\\
\hline
LSTMLM & \textbf{51.30±0.61} &\textbf{41.12±0.64}& \textbf{33.97±0.25} & \textbf{25.17±0.2}9\\
GRULM & \textbf{51.30±0.74} &\textbf{41.00±0.40}& \textbf{33.84±0.34} &\textbf{25.42±0.30}\\
\end{tabular}
\label{tab:results}
\end{table}

\subsection{Results}
\label{sec:results}
We benchmarked eight statistical language models and two neural language models for the task of predicting the next word in radiology reports of \iuxray and \mimic. We randomly sampled reports from \mimic until we obtained a subset with the same number of reports as \iuxray. We additionally removed numbers, punctuation, and turned to lower-case before white space tokenization. We held the 10K last words from each dataset as our test set and used the previous to train our models. Any words occurring less than 10 times were masked with an \textsc{oov} token.

The statistical language models were N-Gram-based models, with factor N varying from 2 (only the previous word was considered) to 9. The neural language models were based either on LSTM or GRU. Following the work of \cite{Spithourakis2016}, we used 50 dimensions for all the hidden representations. Furthermore, we used: a vocabulary of the 1000 most frequent words; a context window of 5 preceding words; uniformly initialized word embeddings of 200 dimensions; a single-layer feed-forward neural network of 100 dimensions and a \textsc{relu} activation before the softmax; Adam optimization and categorical cross entropy; batch size 128; 10\% validation split; early stopping of 100 epochs with patience of 3 epochs; validation loss monitoring.

First, we assessed all models based on their ability to reduce the keystrokes (Keystroke Discounting, $KD$). Since no log files were available for calculating this number directly, we estimated this score based on the length of the words which were correctly predicted by each system. That is, we assume that instead of striking the keyboard as many times as the characters of a word, the physician, during a computer-assisted data entry, simply accepts the correctly predicted word (e.g., by pressing \textsc{tab} or so). More formally, for a sequence of $N$ words ($w^g_1$\dots$w^g_N$) and the respective sequence of system-predicted words ($w^p_1$\dots$w^p_N$), this measure is defined as:
\begin{equation}
\begin{split}
    KD &= 1 - \frac{\sum_{i=1\dots N} dsc(i)}{\sum_{i=1\dots N} |w^g_i|},\\
    dsc(i) &= 
    \begin{cases}
        1, & \text{if } w^g_i = w^p_i\\
        |w^g_i|, & \text{otherwise}
    \end{cases}
    \end{split}
\end{equation}
\noindent
where $|w_i|$ is the number of characters of word $w_i$ and $dsc$ is 1 when the word was correctly predicted by a system and equal to the length of the correct word otherwise. When $KD$ equals to 1, all words are predicted correctly, while when it is equal to 0, no word was predicted correctly. Also, we used micro-averaged accuracy (here, same as precision or recall), which is defined as \emph{the fraction of the correctly predicted words out of all the words in the test}. For both measures the occurrences of the de-identification token (\textsc{`xxxx'}) were disregarded during evaluation, because the ability of the systems to locate candidate de-identification terms is out of the scope of this work. However, in principle, medical language models could be used to assist humans in de-identifying medical texts. And we considered all \textsc{oov} occurrences as system mistakes.

Table~\ref{tab:results} shows the keyword discount (\textsc{KD}) and the micro-averaged accuracy (\textsc{Acc}) scores for the task of next word prediction, for all systems and datasets. Neural language models outperform statistical language models in both datasets by a large margin. In \mimic, the keyword discount is increased by 6 absolute percentage units, from 19.35\% to 25.42\% (or 31\% relative increase). A similar increase was found for accuracy, from 27.34\% to 33.97\% (or 24\% relative increase). In \iuxray, the increase was larger, with 9 absolute percentage units of \textsc{KD} (from 32.30\% to 41.12\%) and 13 absolute percentage units of \textsc{Acc} (from 38.18\% to 51.30\%). The top eight rows of Table~\ref{tab:results} show the different N-Gram based statistical language models. 3-GLM was the best with both evaluation measures in \mimic. In \iuxray, 4-GLM was found as the best in terms of \textsc{Acc} and 5-GLM in \textsc{KD}. We obtained better performance for \iuxray due to its smaller vocabulary size compared to \mimic. 

\begin{figure}
    \centering
    \includegraphics[width=0.48\textwidth]{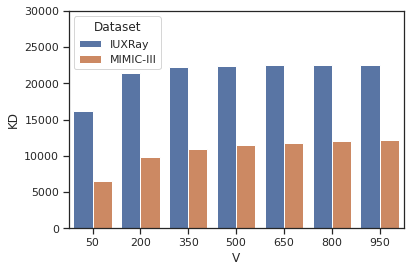}
    \caption{Absolute number of keystroke reduction, by applying LSTMLM only when a frequent vocabulary word is predicted. On the x-axis we see the sizes of the frequent-word sets that are employed.}
    \vspace{-3mm}
    \label{fig:KD_per_V}
\end{figure}

In a real-world setting, however, physicians may prefer to use the advantages of computer-assisted authoring only for specific terms, as for example frequent words, frequent medical terms, or frequent non-medical terms. Thus, in a final experiment, we assumed a deployment setting where the predicted word was only shown if it was one of the frequent ones, and we varied the number of frequent words to be considered. Fig.~\ref{fig:KD_per_V} shows the absolute number of keystrokes omitted when the best performing LSTMLM was applied. Interestingly, even though the target vocabulary is reduced to only 50 words, we can observe a decrease of more than 15K keystrokes. For the case of the 50 most frequent words, without the use of LSTMLM the keystrokes would have been approximately 50K.

\section{Conclusion}
We highlighted the importance of predictive keyboard for medical text and demonstrated the benefits for the physicians in terms of speedups in completing their clinical text reports. Our experimental evaluation on radiology reports from two real-world medical datasets showed that neural language models can achieve an accuracy of up to 51.3\%, which implies that the obtained speedups correspond to a similar factor at the word level. Directions for future work include the investigation of alternative statistical and deep learning models, the consideration of additional medical datasets (e.g., discharge summaries), and measuring the speedups in a real-world application with models deployed in healthcare systems.

\section*{Acknowledgments}
We thank the anonymous reviewers for their comments. This work was supported in part by the Swedish Research Council starting grant Temporal Data Mining for Detective Adverse Events in Healthcare, ref. no. VR-2016-03372 as well as the EXTREME project of the Digital Futures framework.

\bibliographystyle{IEEEtran}
\bibliography{paper}

\end{document}